\def\year{2021}\relax
\documentclass[letterpaper]{article} 
\usepackage{aaai21}  
\usepackage{times}  
\usepackage{helvet} 
\usepackage{courier}  
\usepackage[hyphens]{url}  
\usepackage{graphicx} 
\usepackage{natbib}  
\usepackage{caption} 
\usepackage{multirow}
\usepackage{hyperref}
\usepackage{amsmath}
\usepackage{amssymb}
\usepackage{algorithm}
\usepackage{algorithmic}

\title{Self-supervised Multi-view Stereo via Effective Co-Segmentation and Data-Augmentation}
\author{
    Hongbin Xu\textsuperscript{\rm 1,\rm 3}\thanks{H.Xu and Z.Zhou contributed equally.},
    Zhipeng Zhou\textsuperscript{\rm 1}\footnotemark[1],
    Yu Qiao\textsuperscript{\rm 1,\rm 2}\thanks{Corresponding author.},
    Wenxiong Kang\textsuperscript{\rm 3},
    Qiuxia Wu\textsuperscript{\rm 3}
    \\
}
\affiliations{
    
    \textsuperscript{\rm 1}ShenZhen Key Lab of Computer Vision and Pattern Recognition, \\
    Shenzhen Institutes of Advanced Technology, Chinese Academy of Sciences, Shenzhen, China \\
    \textsuperscript{\rm 2}Shanghai AI Lab, Shanghai, China \\
    \textsuperscript{\rm 3}South China University of Technology, Guangzhou, China \\
    
    hongbinxu1013@gmail.com, \{zp.zhou, yu.qiao\}@siat.ac.cn, \{auwxkang, qxwu\}@scut.edu.cn

}

\begin{document}

\maketitle

\makeatletter 
  \newcommand\figcaption{\def\@captype{figure}\caption} 
  \newcommand\tabcaption{\def\@captype{table}\caption} 
\makeatother

\begin{abstract}
Recent studies have witnessed that self-supervised methods based on view synthesis obtain clear progress on multi-view stereo (MVS).
However, existing methods rely on the assumption that the corresponding points among different views share the same color, which may not always be true in practice.
This may lead to unreliable self-supervised signal and harm the final reconstruction performance.
To address the issue, we propose a framework integrated with more reliable supervision guided by semantic co-segmentation and data-augmentation.
Specially, we excavate mutual semantic from multi-view images to guide the semantic consistency.
And we devise effective data-augmentation mechanism which ensures the transformation robustness by treating the prediction of regular samples as pseudo ground truth to regularize the prediction of augmented samples.
Experimental results on DTU dataset show that our proposed methods achieve the state-of-the-art performance among unsupervised methods, and even compete on par with supervised methods.
Furthermore, extensive experiments on Tanks\&Temples dataset demonstrate the effective generalization ability of the proposed method.
\end{abstract}

\section{Introduction}
\label{sec:introduction}

Multi-view stereo (MVS) aims at recovering 3D scenes from multi-view images and calibrated cameras, which is an important problem and widely studied in computer vision community \cite{seitz2006comparison}.
Recent success of deep learning has triggered the interest of extending MVS pipelines to end-to-end neural networks.
The learning-based methods \cite{yao2018mvsnet, yao2019recurrent} adopt CNNs to estimate the feature maps and build a cost volume upon the reference camera frustum to predict a per-view depth map for reconstruction.
With the help of large-scale 3D ground truth, they outperform traditional geometry-based approaches and dominate the leaderboard.
Whereas the learning-driven approaches strongly depend on the availability of 3D ground truth data for training, which is not easy to acquire \cite{zhong2018open}.
Thus it drives the community to focus on unsupervised/self-supervised MVS approaches.

\begin{figure}[t]
    \centering
    \includegraphics[width=0.9\columnwidth]{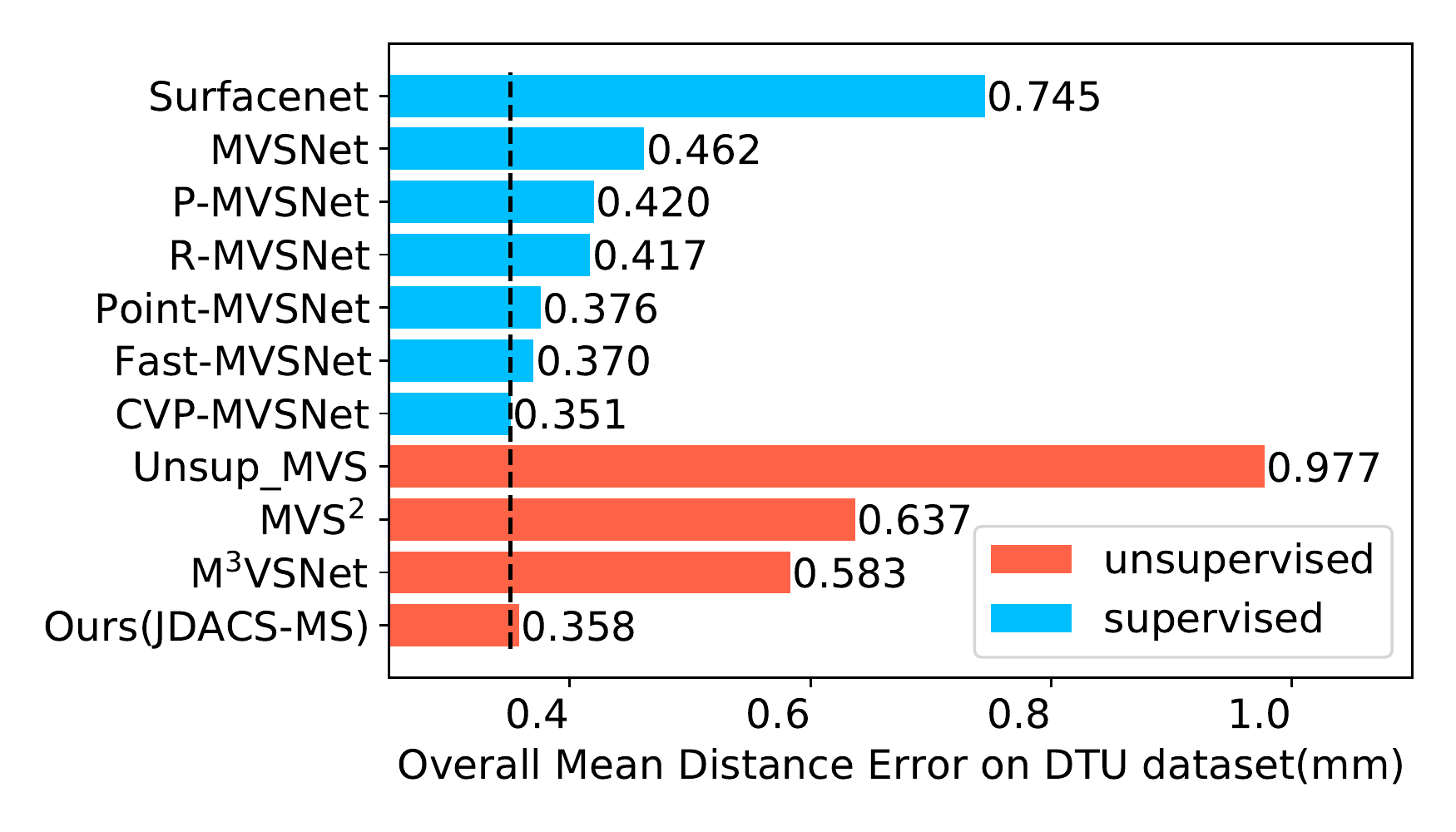}
    \caption{Comparison between SOTA supervised and unsupervised MVS methods.}
    \label{fig1}
\end{figure}

Recently, there has been a surge in the number of self-supervised MVS methods that transform the depth estimation problem to an image reconstruction problem \cite{khot2019learning, dai2019mvs2, huang2020m}.
The predicted depth map and the input image are used to reconstruct the image on another view, thus the self-supervision loss is built to estimate the difference between the reconstructed and realistic image on that view.
However, as summarized in Figure \ref{fig1}, despite the impressive efforts in previous unsupervised methods, there still exists a clear gap between supervised and unsupervised results.
In this paper, we suggest to rethink the task of self-supervision itself to improve the accuracy in MVS.

\begin{figure*}[t]
    \centering
    \includegraphics[width=0.9\textwidth]{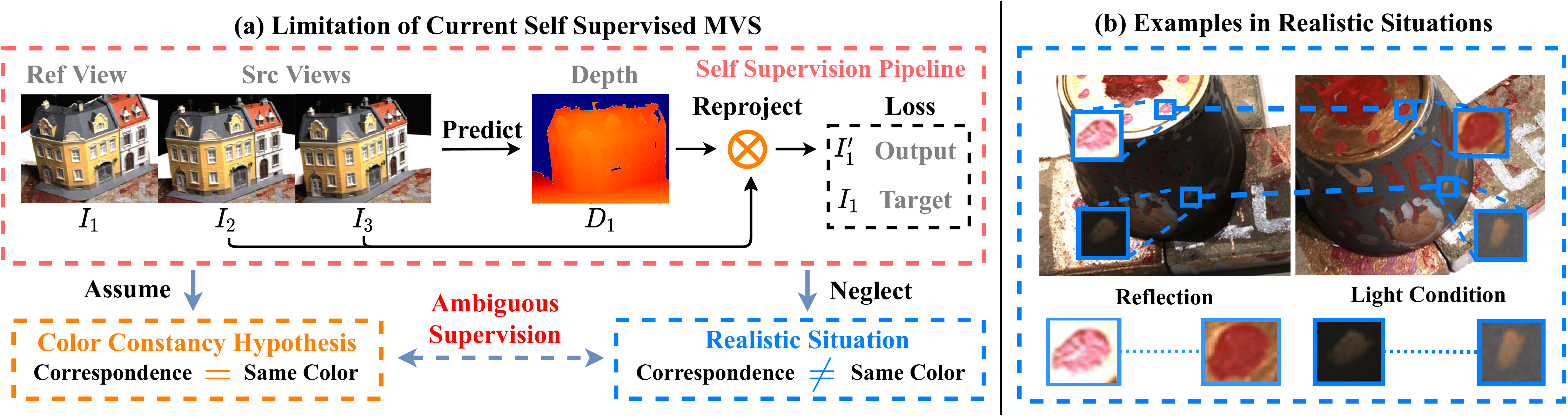}
    \caption{Illustration of the color constancy ambiguity problem in self-supervised MVS.}
    \label{fig2}
\end{figure*}

Previous self-supervised MVS methods largely rely on the same \textit{color constancy hypothesis}, assuming the corresponding points among different views have the same color.
However, as Figure \ref{fig2} shows, in realistic scenarios, various factors may disturb the color distribution, such as light conditions, reflections, noise, etc.
Consequently, the ideal self-supervision loss is susceptible to be confused by these common disturbances in color, leading to ambiguous supervision in challenging scenarios, namely \emph{color constancy ambiguity}.
To address the issues, we aim to incorporate  the following extra priors of correspondence with the prior of color constancy in self-supervision loss:
(1) \emph{The prior of semantic correspondence can provide abstract matching clues to guide the supervision}.
(2) \emph{The prior of data augmentation consistency can enhance the robustness towards color fluctuation}.
Hence, we propose a novel Joint Data-Augmentation and Co-Segmentation self-supervised MVS framework, namely JDACS.

For the prior of semantic consistency, most of the previous methods rely on the manually annotated semantic labels \cite{yang2018segstereo, dovesi2019real} restricted in fixed scenarios like autonomous driving with specified semantic classes.
Whereas in the concern of MVS, on the one hand the semantic annotations are relatively expensive, on the other hand the huge variation in scenarios makes the semantic categories unfixed for segmentation which requires specified classes.
Differently, we adopt non-negative matrix factorization (NMF) \cite{ding2005equivalence} to excavate the common semantic clusters among multi-view images dynamically for unsupervised co-segmentation \cite{collins2018deep}.
Then the semantic consistency is maximized among the re-projected multi-view semantic maps.

For the prior of data augmentation consistency, heavy data augmentation seldom appears in previous self-supervised MVS methods \cite{khot2019learning, dai2019mvs2, huang2020m}, because the natural color fluctuation in data augmentation will lead to the color constancy ambiguity in self-supervision.
To preserve the reliability of self-supervision, we attach an additional data-augmentation branch with various transformations to the regular training branch.
The output of regular training branch is taken as pseudo ground truth to supervise the output of augmented training branch.

In summary, our contributions are: 

(1) We propose a unified unsupervised MVS pipeline called Joint Data-Augmentation and Co-Segmentation framework(JDACS) where extra priors of semantic consistency and data augmentation consistency can provide reliable guidance to overcome the color constancy ambiguity. 

(2) We propose a novel self-supervision signal based on semantic consistency, which can excavate mutual semantic correspondences from multi-view images at unfixed scenarios in a totally unsupervised manner. 

(3) We propose a novel way to incorporate heavy data augmentation into unsupervised MVS, which can provide regularization towards color fluctuation. 

(4) The experimental results show that our proposed method can lead to a leap of performance among unsupervised methods and compete on par with some top supervised methods.


\begin{figure*}[t]
    \centering
    \includegraphics[width=0.9\textwidth]{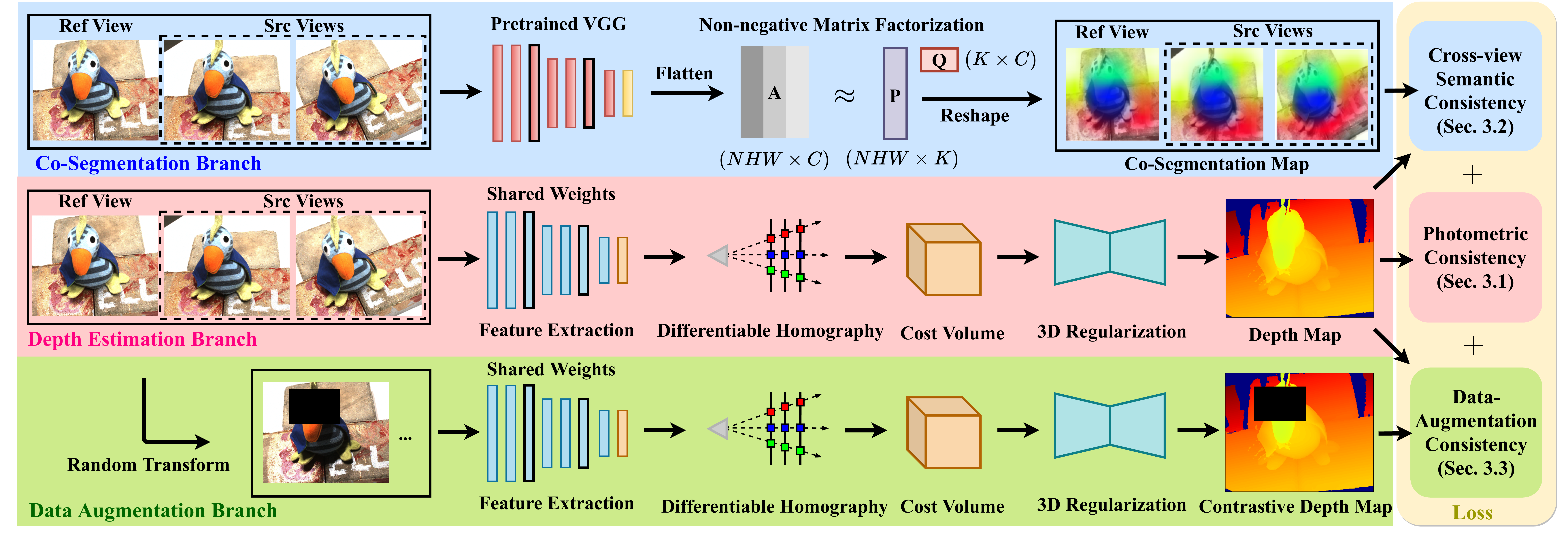}
    \caption{Illustration of our Joint Data-Augmentation and Co-Segmentation (JDACS) MVS framework.}
    \label{fig3}
\end{figure*}

\section{Related Work}

\textbf{Supervised MVS:}
Recent advances in deep learning have interested a series of learnable systems for solving MVS problems \cite{huang2018deepmvs, ji2017surfacenet}.
MVSNet \cite{yao2018mvsnet} is an end-to-end MVS pipeline that builds a cost volume upon the reference camera frustum and learns the 3D regularization with CNNs.
Many variants based on MVSNet have been proposed for improving the performance \cite{yao2019recurrent, luo2019p}.
Concurrently, along with the fervor for expanding the MVS framework to a coarse-to-fine manner, \cite{chen2019point, yu2020fast, yang2020cost, cheng2020deep, gu2019cas, xu2020learning} separate the single MVS pipeline into multiple stages, achieving impressive performances.

\noindent\textbf{Unsupervised MVS:}
Under the assumption of photometric consistency \cite{godard2017unsupervised}, unsupervised learning has been developed in multi-view systems.
\cite{khot2019learning} inherit the self-supervision signal based on view synthesis and dynamically aggregates informative clues from nearby views.
\cite{dai2019mvs2} predict the depth maps for all views simultaneously and filter the occluded regions.
\cite{huang2020m} further endow the depth-normal consistency into the MVS pipeline for improvement.
Whereas all these methods share the assumption of color constancy, suffering from ambiguous supervision in challenging scenarios.

\noindent\textbf{Segmentation Guided Algorithms:}
By assigning each pixel in the image to a specific class, semantic segmentation \cite{long2015fully} can provide an abstract representation.
Several methods incorporate the scene parsing information with other tasks.
SegStereo \cite{yang2018segstereo} enables joint learning for segmentation and disparity esitimation simultaneously and \cite{cheng2017segflow} utilize semantic clues to guide the training of optical flow estimation.
These methods rely on annotated labels for segmentation in specific scenes like autonomous driving, whereas we differently concentrate on excavating semantics from dynamic scenarios.
Co-segmentation methods aim at predicting foreground pixels of objects given an image collection \cite{joulin2012multi}.
We apply unsupervised co-segmentation \cite{casser2019unsupervised} on the multi-view pairs to exploit the common semantics.


\section{Method}

In this section, we present Joint Data-Augmentation and Co-Segmentation framework(JDACS).
To improve the reliability towards color constancy ambiguity, we incorporate extra priors of semantic consistency and data-augmentation consistency with a basic structure of deep MVS pipeline \cite{yao2018mvsnet} in JDACS.
As Figure \ref{fig3} shows, the architecture of JDACS consists of Depth Estimation branch, Co-Segmentation branch and Data-Augmentation branch. 

\subsection{Depth Estimation Branch}
\label{depth-estimation}

As an unsupervised method, our proposed framework can be combined with arbitrary MVS networks.
Here, we adopt MVSNet \cite{yao2018mvsnet} as a representative backbone.
The network firstly extracts features using a CNN from $N$ input images.
Then a variance-based cost volume is constructed via differentiable homography warping and a 3D U-Net is used to regularize the 3D cost volume.
Finally, the depth map is inferred for every reference image.
A sketch of the pipeline is shown in Figure \ref{fig3}.

\noindent\textbf{Photometric Consistency:}
The key idea of photometric consistency \cite{barnes2009patchmatch} is to minimize the difference between synthesized image and original image on the same view.
Denote that the $1$-st view is the reference view and the remaining $N-1$ views as source views indexed by $i(2 \leq i \leq N)$.
For a particular pair of images $(I_1, I_i)$ with associated intrinsic and extrinsic parameters $(K, T)$.
We can calculate the corresponding position $p'_j$ in source view based on its coordinate $p_j$ in reference view.
\begin{equation}
    p'_j = KT(D(p_j)K^{-1}p_j)
    \label{eq1}
\end{equation}
where $j(1 \leq j \leq HW)$ is the index of pixels and $D$ represents the predicted depth map.

The warped image $I'_i$ can then be obtained by using the differentiable bilinear sampling from $I_i$.
\begin{equation}
    I'_i(p_j) = I_i(p'_j)
    \label{eq2}
\end{equation}

Along with the warping, a binary validity mask $M_i$ is generated simultaneously, indicating valid pixels in the novel view because some pixels may be projected to the external area of images.
In a MVS system, we can warp all $N-1$ source views to the reference view to calculate the loss.
\begin{equation}
    L_{PC} = \sum\limits_{i=2}^{N} \frac{|| (I'_i - I_1)\odot M_i ||_2 + || (\nabla I'_i - \nabla I_1)\odot M_i ||_2 }{||M_i||_1}
    \label{eq3}
\end{equation}
where $\nabla$ denotes the gradient operator and $\odot$ is dot product.

\subsection{Co-Segmentation Branch}
In previous methods \cite{yang2018segstereo, casser2019unsupervised}, handcrafted semantic annotations are usually utilized to provide extra supervision to improve the performance.
However, due to the huge variation of scenarios and the expensive cost for manual annotations in MVS, we differently choose to mine the implicit common segments from multi-view images via unsupervised co-segmentation. 
Co-segmentation aims at localizing the foreground pixels of the common objects given an image collection.
It has been proven that non-negative matrix factorization (NMF) has an inherent clustering property in \cite{ding2005equivalence}.
Following a classical co-segmentation pipeline \cite{collins2018deep}, NMF applied to the activations of a pretrained CNN layer can be exploited to find semantic correspondences across images.

\noindent\textbf{Non-negative Matrix Factorization:}
Non-negative matrix factorization(NMF) is a group of algorithms in multivariate analysis and linear algebra where a matrix $A$ is factorized into two matrices $P$ and $Q$. 
All the three matrices are with the property that having no negative elements. 
As \cite{ding2005equivalence} shows, NMF has an inherent clustering property that it automatically clusters the columns of matrix $A=(a_1,...,a_n)$. 
More specifically, if we impose an orthonormal constraint on $Q$($QQ^{T}=I$), then the approximation of $A$ by $A\simeq PQ$ achieved by minimizing the following error function is equivalent to the optimization of K-means clustering.
\begin{equation}
    ||A-PQ||_{F}, P\geq 0,Q\geq 0
    \label{eq4}
\end{equation}
where the subscript $F$ means the Frobenius Norm. 

\begin{figure}[t]
    \centering
    \includegraphics[width=0.9\columnwidth]{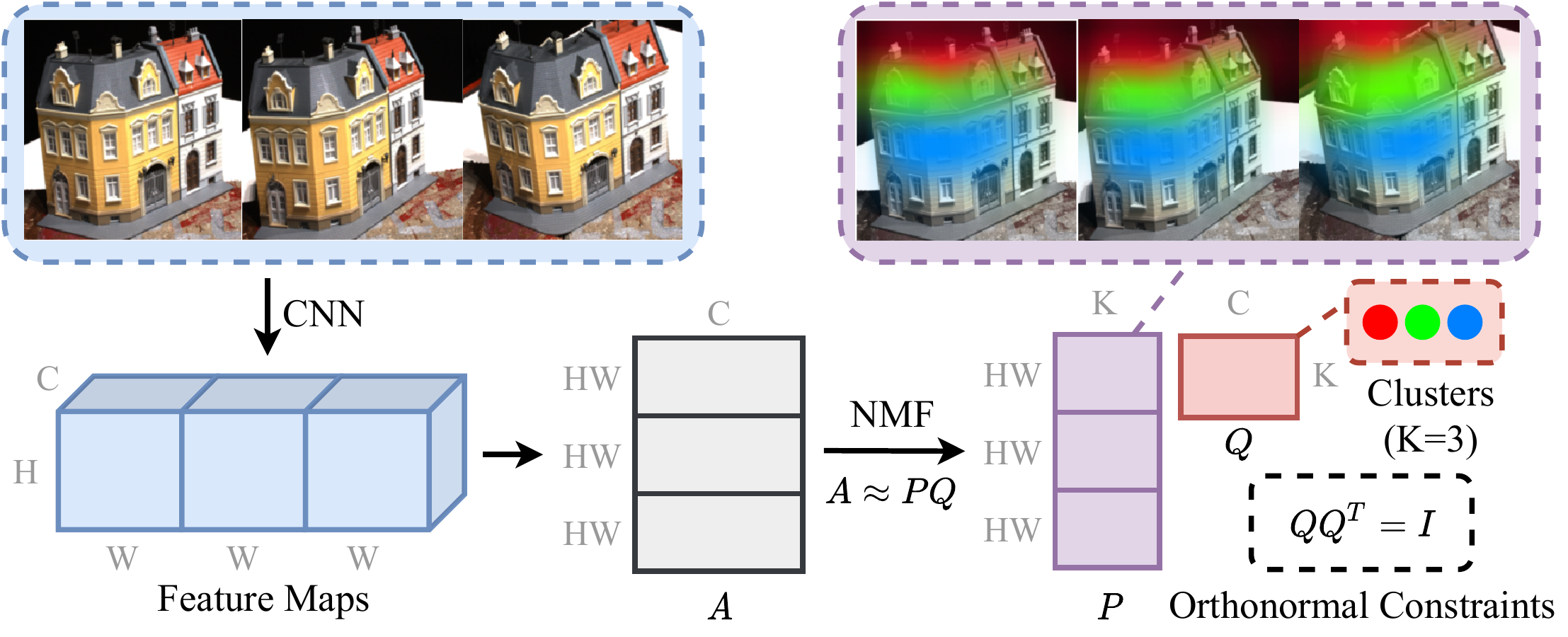}
    \caption{Brief illustration of the clustering effect of NMF.}
    \label{fig-nmf}
\end{figure}

\noindent\textbf{Clustering on CNN Activations:}
ReLU is a common component for many modern CNNs, due to its desirable gradient properties.
The CNN feature maps activated by ReLU result in non-negative activations, which naturally fit for the target of NMF.
As shown in Figure \ref{fig3}, we apply a pretrained VGG network \cite{simonyan2014very} for feature extraction.
Denote that the extracted feature map is of dimension $(H, W, C)$ on each of the $N$ views.
Then the multi-view feature maps are concatenated and reshaped to a $(NHW,C)$ matrix $A$.
By utilizing multiplicative update rule in \cite{ding2005equivalence} to solve NMF, $A$ is factorized into a $(NHW,K)$ matrix $P$ and $(K,C)$ matrix $Q$, where $K$ is the NMF factors representing the number of semantic clusters.
For a comprehensive understanding, we provide a brief interpretation of the results $P$, $Q$ and the clustering effect of NMF in Figure \ref{fig-nmf}.

\noindent\textbf{The Q matrix:} 
Due to the orthonormal constraints of NMF($Q Q^T = I$) \cite{ding2005equivalence}, each row of the $(K,C)$ matrix Q can be viewed as a cluster centroid of $C$ dimensions, which corresponds to a coherent object among views.

\noindent\textbf{The P matrix:}
The rows of the $(NHW,K)$ matrix P correspond to the spatial positions of all pixels from $N$ views. 
In general, the matrix factorization $A \approx P Q$ enforces the product between each row of $P$ and each column of $Q$ to best approximate the $C$ dimensional feature of each pixel in $A$.
As shown in Figure \ref{fig-nmf}, $K=3$ semantic objects are clustered in $Q$ from the feature embeddings of all pixels in $A$, thus $P$ contains the similarity between each pixel and each of the $K=3$ clustered semantic objects.
Consequently, $P$ can further be reshaped into $N$ heat maps of dimension $(H, W, K)$ and fed into a softmax layer to construct the co-segmentation maps $S$.

\noindent\textbf{Semantic Consistency Loss:}
With the co-segmentation maps $S$ extracted from matrix $P$, we can design a self-supervision constraint based on semantic consistency.
The key idea is to expand the photometric consistency across multiple views \cite{barnes2009patchmatch} to the segmentation maps.
Similar to the photometric consistency discussed in Section \ref{depth-estimation}, we can calculate the corresponding position $p'_j$ in source views with the pixel $p_j$ in reference view according to Equation \ref{eq1}, given the predicted depth value $D(p_j)$ and the $j$-th pixel in the image.
Then the warped segmentation map $S'_i$ from the $i$-th source view can be reconstructed by bilinear sampling.
\begin{equation}
    S'_i(p_j) = S_i(p'_j)
    \label{eq5}
\end{equation}

Finally, the semantic-consistency objective $L_{SC}$ is measured by calculating the per-pixel cross-entropy loss between the warped segmentation map $S'_i$  and the ground truth labels converted from reference segmentation map $S_1$.
\begin{equation}
    L_{SC}=-\sum\limits_{i=2}^{N} [\frac{1}{||M_i||_1} \sum\limits_{j=1}^{HW} f (S_{1, j}) \log (S'_{i, j}) M_{i,j}]
    \label{eq6}
\end{equation}
where $f(S_{1, j}) = \mathop{onehot} (\mathop{\arg\max} (S_{1, j}))$ and $M_i$ is a binary mask indicating valid pixels from the $i$-th view to reference view.

\subsection{Data-Augmentation Branch}

Some recent works \cite{xie2019unsupervised, chen2020simple} in contrastive learning demonstrate the benefits of data augmentation in self-supervised learning.
The intuition is that data augmentation brings challenging samples which bust the reliability of unsupervised loss and hence provides robustness towards variations.

Briefly, a random vector $\theta$ is defined to parameterize an arbitrary augmentation $\tau_{\theta}:I \rightarrow \bar{I}_{\tau_{\theta}}$ on image $I$. 
However, data augmentation has seldom been applied in self-supervised methods \cite{khot2019learning, dai2019mvs2, huang2020m}, because natural color fluctuation in augmented images may disturb the color constancy constraint of self-supervision.
Hence, we enforce the unsupervised data augmentation consistency by contrasting the output of original data and augmented samples as a regularization, instead of optimizing the original objective of view synthesis.

\noindent\textbf{Data Augmentation Consistency Loss:}
Specifically, as shown in Figure \ref{fig3}, the prediction of a regular forward pass for original images $I$ in Depth Estimation branch is denoted as $D$.
Accordingly, the prediction of augmented images $\bar{I}_{\tau_{\theta}}$ is denote as $\bar{D}_{\tau_{\theta}}$.
In a contrastive manner, the data-augmentation consistency is ensured by minimizing the difference between $D$ and $\bar{D}_{\tau_{\theta}}$:
\begin{equation}
    L_{DA} = \frac{1}{||M_{\tau_{\theta}}||_1} \sum ||(D - \bar{D}_{\tau_{\theta}})\odot M_{\tau_{\theta}}||_2
    \label{eq7}
\end{equation}
where $M_{\tau_{\theta}}$ represents the unoccluded mask under transformation $\tau_{\theta}$.
Due to the epipolar constraints among different views, the integrated augmentation methods in our framework should not change the spatial location of pixels.
We will show some augmentation methods used in our method as follows:

\noindent\textbf{Cross-view Masking:}
To simulate the occlusion hallucination among the multi-view situations, we randomly generate a binary crop mask $1-M_{\tau_{\theta_1}}$ to block out some regions on reference view.
Then the occlusion mask is projected to other views to mask out the corresponding area in images.
Following the assumption that the remaining regions $M_{\tau_{\theta_1}}$ should be immune to the transformation, we can contrast the validity regions between the results of original and augmented samples.

\noindent\textbf{Gamma Correction:}
Gamma correction is a nonlinear operation used to adjust the illuminance of images.
To simulate various illuminations, we integrate random gamma correction $\tau_{\theta_2}$ parameterized by $\theta_2$ to challenge the unsupervised loss.

\noindent\textbf{Color Jitter and Blur:}
Many transformations can attach color fluctuation to images, such as random color jitter, random blur, random noise.
The color fluctuation makes the unsupervised loss in MVS unreliable, because the photometric loss requires the color constancy among views.
In contrast, these transformations denoted as $\tau_{\theta_3}$ can create challenging scenes and regularize the robustness towards color fluctuation in self-supervision.

The overall transformation $\tau_{\theta}$ can be represented as a combination of the aforementioned augmentations: $\tau_\theta=\tau_{\theta_3}\circ\tau_{\theta_{2}}\circ\tau_{\theta_1}$, where $\circ$ represents function composition.

\begin{table}[t]
\centering
\small
\smallskip\begin{tabular}{c|c|ccc}
\hline 
 & Method & Acc. & Comp. & Overall\tabularnewline
\hline 
\hline 
\multirow{4}{*}{Geo.} & Furu & 0.613 & 0.941 & 0.777\tabularnewline
 & Tola & 0.342 & 1.190 & 0.766\tabularnewline
 & Camp & 0.835 & 0.554 & 0.694\tabularnewline
 & Gipuma & 0.283 & 0.873 & 0.578\tabularnewline
\hline 
\multirow{7}{*}{Sup.} & Surfacenet & 0.450 & 1.040 & 0.745\tabularnewline
 & MVSNet & 0.396 & 0.527 & 0.462\tabularnewline
 & P-MVSNet & 0.406 & 0.434 & 0.420\tabularnewline
 & R-MVSNet & 0.383 & 0.452 & 0.417\tabularnewline
 & Point-MVSNet & 0.342 & 0.411 & 0.376\tabularnewline
 & Fast-MVSNet & 0.336 & 0.403 & 0.370\tabularnewline
 & CVP-MVSNet & 0.296 & 0.406 & 0.351\tabularnewline
\hline 
\multirow{5}{*}{UnSup.} & Unsup\_MVS & 0.881 & 1.073 & 0.977\tabularnewline
 & MVS$^{2}$ & 0.760 & 0.515 & 0.637\tabularnewline
 & M$^{3}$VSNet & 0.636 & 0.531 & 0.583\tabularnewline
 & \textbf{JDACS} & \textbf{0.571} & \textbf{0.515} & \textbf{0.543}\tabularnewline
 & \textbf{JDACS-MS} & \textbf{0.398} & \textbf{0.318} & \textbf{0.358}\tabularnewline
\hline 
\end{tabular}
\caption{Quantitative results on DTU evaluation benchmark. Geo. represents traditional geometric methods. Sup. represents supervised methods. UnSup. represents unsupervised methods.}\smallskip
\label{table1}
\end{table}

\begin{figure}[t]
    \centering
    \includegraphics[width=0.9\columnwidth]{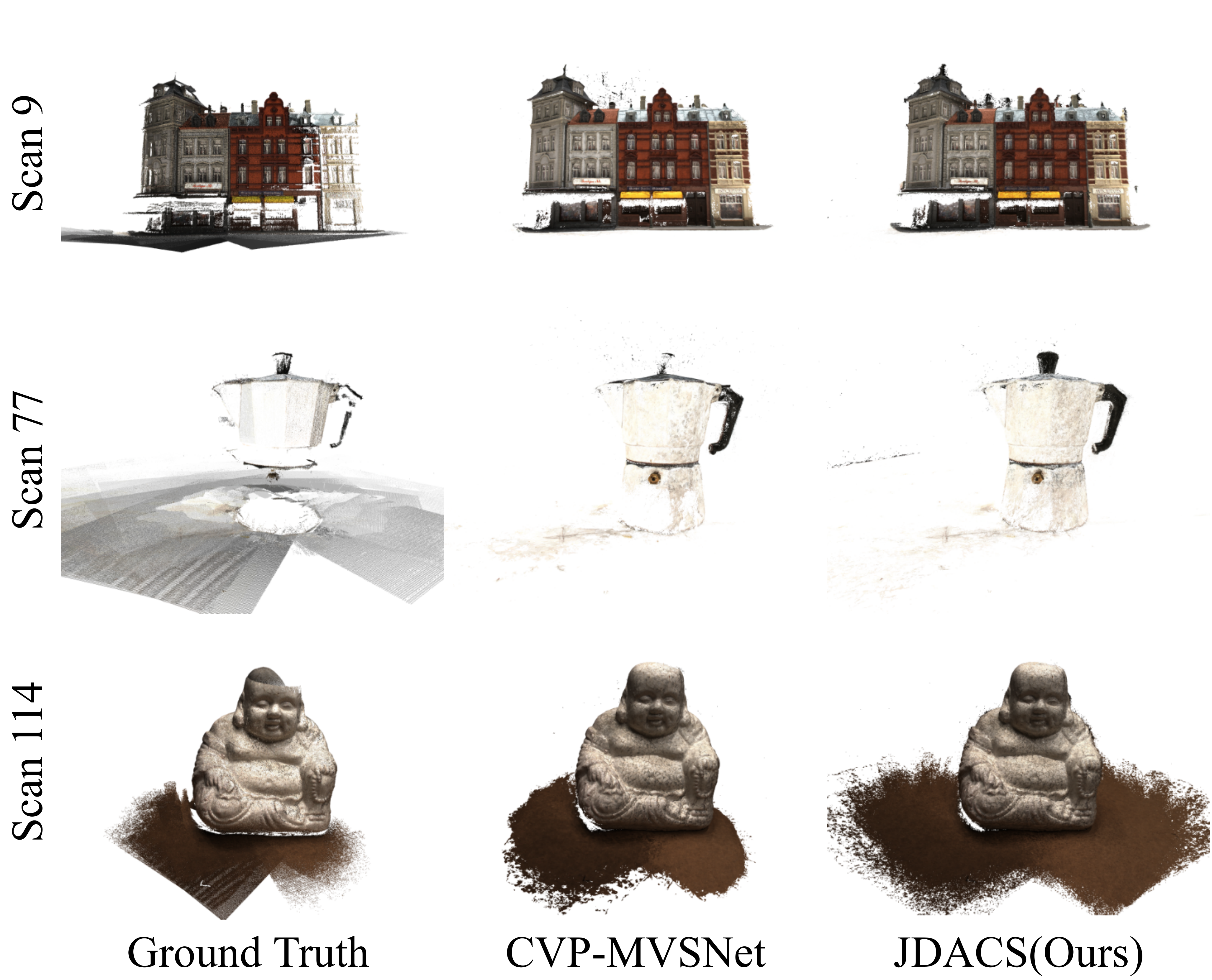}
    \caption{Qualitative comparison in 3D reconstruction between our JDACS and SOTA supervised method(CVP-MVSNet) on DTU dataset. From left to right: ground truth, results of supervised CVP-MVSNet, our results.}
    \label{fig4}
\end{figure}

\subsection{Overall Architecture and Loss}

As shown in Figure \ref{fig3}, the overall framework has three components: Depth Estimation branch, Co-Segmentation branch and Data-Augmentation branch.
In our paper, we aim to handle the color constancy ambiguity problem in self-supervised MVS, as discussed in Section \ref{sec:introduction}.
Apart from the basic self-supervision signal based on photometric consistency $L_{PC}$ (Equation \ref{eq1}), we add two extra self-supervision signals of semantic consistency $L_{SC}$ and data-augmentation consistency $L_{DA}$ to the framework.
In addition to the aforementioned loss, some common regularization terms suggested by \cite{mahjourian2018unsupervised, khot2019learning} for depth estimation are applied, such as structured similarity $L_{SSIM}$ and depth smoothness $L_{Smooth}$.

\begin{table*}[t]
\centering
\small
\begin{tabular}{c|ccc|ccc}
\hline 
Method & Supervised & Input Size & Depth Map Size & Acc. & Comp. & Overall \\ \hline \hline 
MVSNet & $\checkmark$ & $1152\times864$ & $288\times216$ & 0.456 & 0.646 & 0.551 \\
JDACS & $\times$ & $1152\times864$ & $288\times216$ & 0.571 & 0.515 & 0.543 \\ \hline 
CVP-MVSNet & $\checkmark$ & $1600\times1152$ & $1600\times1152$ & 0.296 & 0.406 & 0.351 \\
JDACS-MS & $\times$ & $1600\times1152$ & $1600\times1152$ & 0.398 & 0.318 & 0.358 \\
\hline 
\end{tabular}
\caption{Comparison between the backbone networks with same settings trained by supervision and our JDACS self-supervision framework. Due to the GPU memory limitation, we decrease the resolution of MVSNet to $1152 \times 864$ as \cite{chen2019point}.}\smallskip
\label{table2}
\end{table*}

\begin{figure*}[t]
    \centering
    \includegraphics[width=0.9\textwidth]{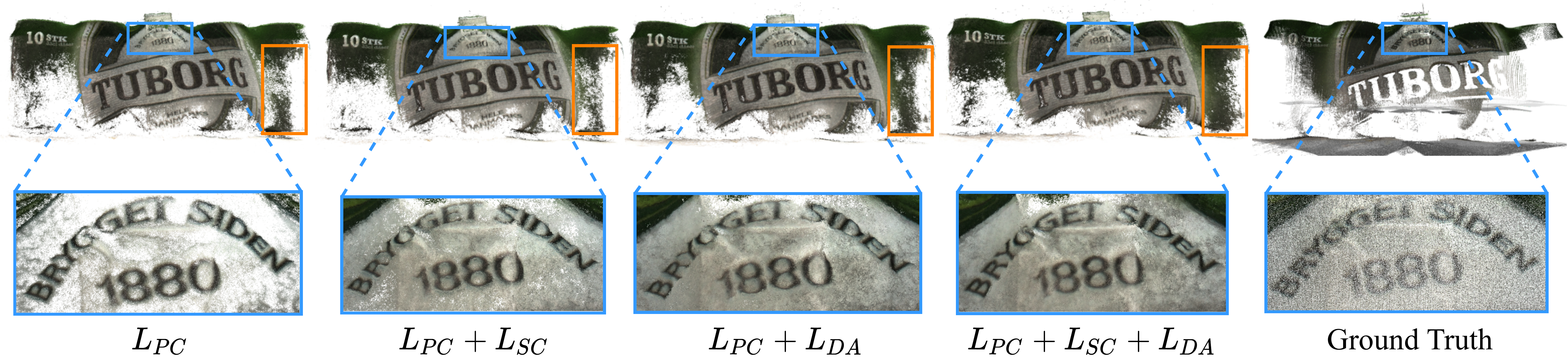}
    \caption{Qualitative results JDACS on \emph{scan12} of the DTU dataset. Top row: Overview of generated point clouds with different combinations of self-supervision components. Bottom row: zoomed local areas. $L_{PC}$: Photometric-Consistency Loss; $L_{SC}$: Semantic-Consistency Loss; $L_{DA}$: Data-Augmentation-Consistency Loss.}
    \label{fig5}
\end{figure*}

\begin{table}[t]
\centering
\small
\begin{tabular}{ccc|ccc}
\hline 
 $L_{PC}$ & $L_{SC}$ & $L_{DA}$ & Acc. & Comp. & Overall\tabularnewline
    \hline 
    \hline 
     $\checkmark$ &  &   & 0.7215 & 0.6339 & 0.6777\tabularnewline
     $\checkmark$ & $\checkmark$ &   & 0.6134 & 0.5771 & 0.5953\tabularnewline
     $\checkmark$ &  & $\checkmark$ & 0.5908 & 0.5887 & 0.5898\tabularnewline
     $\checkmark$ & $\checkmark$ & $\checkmark$ & \textbf{0.5713} & \textbf{0.5146} & \textbf{0.5429}\tabularnewline
    \hline 
\end{tabular}
\caption{Ablation Study of different components in our JDACS self-supervision network.}
\label{table3}
\end{table}

\begin{table}[t]
\centering
\small
\begin{tabular}{ccc|ccc}
    \hline 
    $L_{PC}$ & $L_{SC}$ & $L_{DA}$ & Acc. & Comp. & Overall\tabularnewline
    \hline 
    \hline 
     $\checkmark$ &  &  & 0.4645 & 0.4092 & 0.4369\tabularnewline
     $\checkmark$ & $\checkmark$ &  & 0.4433 & 0.3892 & 0.4163\tabularnewline
     $\checkmark$ &  & $\checkmark$ & 0.4330 & 0.3373 & 0.3851\tabularnewline
     $\checkmark$ & $\checkmark$ & $\checkmark$ & \textbf{0.3977} & \textbf{0.3177} & \textbf{0.3577}\tabularnewline
    \hline 
\end{tabular}
\caption{Ablation Study of different components in our JDACS-MS self-supervision network.}
\label{table4}
\end{table}

The final objective can be constructed as follows:
\begin{equation}
\begin{split}
    L = \lambda_1 L_{PC} + \lambda_2 L_{SC} + \lambda_3 L_{DA} \\ 
    + \lambda_4 L_{SSIM} + \lambda_5 L_{Smooth}
\end{split}
\label{eq8}
\end{equation}
where the weights are empirically set as: $\lambda_1=0.8$, $\lambda_2=0.1$, $\lambda_3=0.1$, $\lambda_4=0.2$, $\lambda_5=0.0067$.

\section{Experiments}

In this section, we conduct comprehensive experiments to evaluate the proposed JDACS framework.
First, we introduce the implementation details.
Then, we evaluate the proposed method on \emph{DTU benchmark} \cite{aanaes2016large} and further conduct ablation studies to analyze the significant components.
At last, we test the proposed method on \emph{Tanks\&Temples benchmark} \cite{knapitsch2017tanks} to verify the generalization ability.





\subsection{Implementation Details}

\begin{table}[t]
\centering
\small
\smallskip\begin{tabular}{c|ccc}
\hline 
Clusters & Acc. & Comp. & Overall\tabularnewline
\hline 
\hline 
$K=2$ & 0.6166 & 0.5752 & 0.5959\tabularnewline
$K=4$ & \textbf{0.6134} & \textbf{0.5771} & \textbf{0.5953}\tabularnewline
$K=6$ & 0.6207 & 0.5827 & 0.6017\tabularnewline
$K=8$ & 0.6224 & 0.6030 & 0.6127\tabularnewline
\hline 
\end{tabular}
\caption{Ablation Study of different numbers of semantic clusters $K$.}
\label{table5}
\end{table}

\begin{figure}[t]
\centering
\includegraphics[width=0.9\columnwidth]{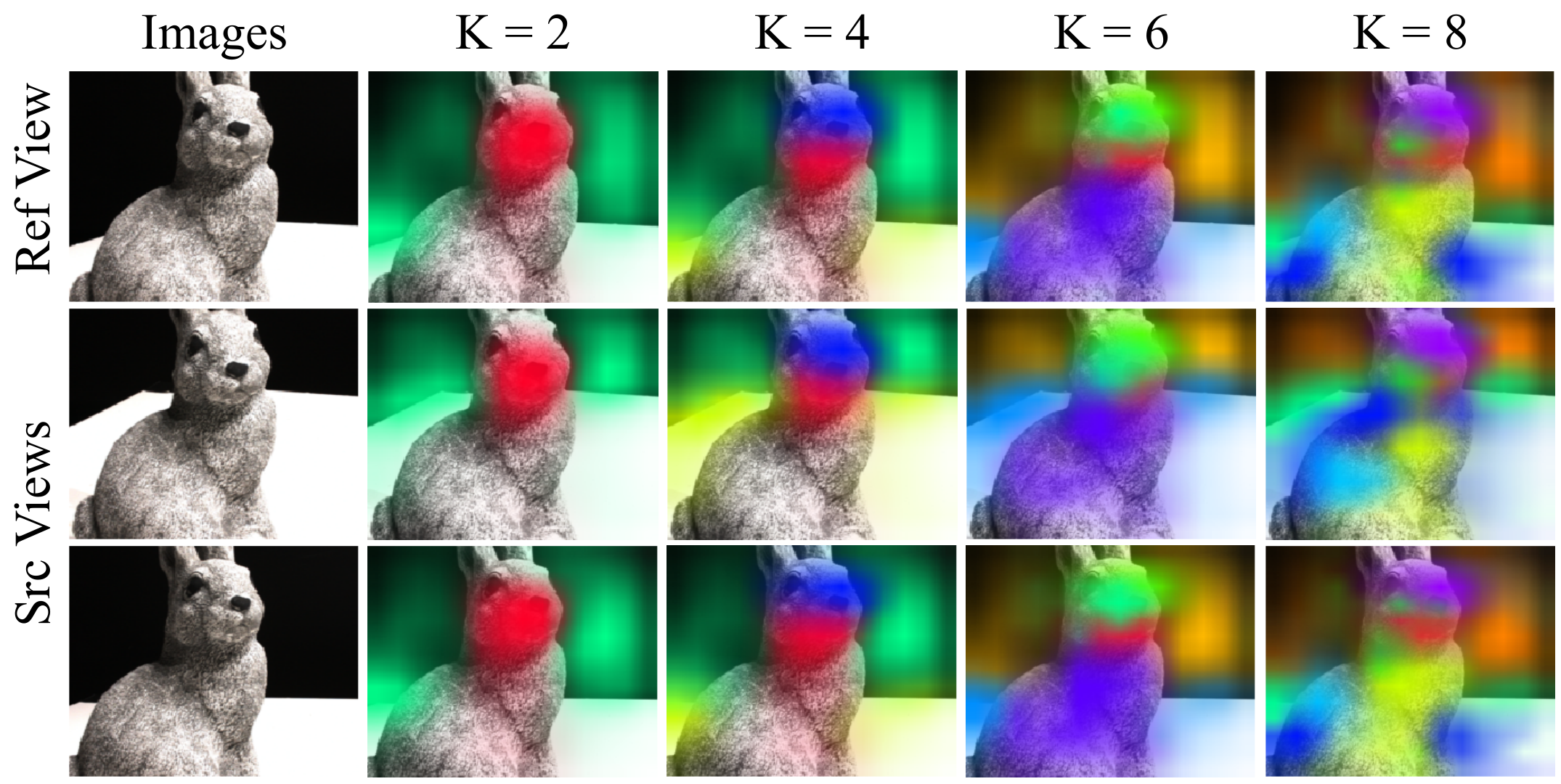} 
\caption{Visualization of the co-segmentation results with different number of segmentation parts $K$.}
\label{fig6}
\end{figure}

\begin{table*}[t]
\centering
\small
\begin{tabular}{c|c|cccccccc}
\hline 
Method & Mean & Family & Francis & Horse & Lighthouse & M60 & Panther & Playground & Train\tabularnewline
\hline 
\hline 
MVS$^{2}$ & 37.21 & 47.74 & 21.55 & 19.50 & 44.54 & 44.86 & \textbf{46.32} & 43.38 & 29.72\tabularnewline
\hline 
M$^{3}$VSNet & 37.67 & 47.74 & 24.38 & 18.74 & 44.42 & 43.45 & 44.95 & 47.39 & 30.31\tabularnewline
\hline 
 Ours & \textbf{45.48} & \textbf{66.62} & \textbf{38.25} & \textbf{36.11} & \textbf{46.12} & \textbf{46.66} & 45.25 & \textbf{47.69} & \textbf{37.16}
 \tabularnewline
\hline 
\end{tabular}
\caption{Quantitative comparison with previous unsupervised methods without finetuning on Tanks\&Temples dataset.}
\label{table6}
\end{table*}

\noindent\textbf{Backbone:}
In default, the most concise MVSNet \cite{yao2018mvsnet} is applied as backbone in our JDACS framework.
We denote the framework as JDACS-MS if a multi-stage MVSNet like CVP-MVSNet \cite{yang2020cost} is selected as backbone.

\noindent\textbf{Training and Testing:}
During the training phase, we only use the training set of DTU without any ground truth depth maps.
Our proposed JDACS\footnote{The code is released at: \url{https://github.com/ToughStoneX/Self-Supervised-MVS}} is implemented in Pytorch and trained on 4 NVIDIA RTX 2080Ti GPUs.
In default, the hyper-parameters during training and testing phase follow the same setting of Unsup\_MVS \cite{khot2019learning}.
With a pattern of data-parallel, the batch size is set to 1 per GPU for JDACS and 4 per GPU for JDACS-MS, which consume no more than 10G memories in each GPU.
We use Adam optimizer with a learning rate of 0.001 which decreases by 0.5 times for every two epochs.
JDACS is trained for 10 epochs as MVSNet \cite{yao2018mvsnet} and JDACS-MS is trained for 27 epochs as CVP-MVSNet\cite{yang2020cost}.

\noindent\textbf{Error Metrics:}
In the DTU benchmark, \emph{Accuracy} is measured as the distance from the result to the ground truth, encapsulating the quality of reconstruction;
\emph{Completeness} is measured as the distance from the ground truth to the result, encapsulating how much of the surface is captured;
\emph{Overall} is a the average of \emph{Accuracy} and \emph{Completeness}, acting as a compositive error metric.
In the Tanks\&Temples benchmark, \emph{F-score} in each scene is calculated following the official evaluation process.



\subsection{Benchmark Results on DTU}

\noindent\textbf{Comparison with SOTA:}
The official metrics of the DTU dataset \cite{aanaes2016large} are: \emph{Accuracy}, \emph{Completeness} and \emph{Overall}.
These metrics are used to compare our proposed methods with other methods.
The comparison includes traditional methods such as Furu \cite{furukawa2009accurate}, Tola \cite{tola2012efficient}, Camp \cite{campbell2008using}, Gipuma \cite{galliani2015massively}.
For the supervised methods, single stage networks such as Surfacenet \cite{ji2017surfacenet}, MVSNet \cite{yao2018mvsnet}, P-MVSNet \cite{luo2019p}, R-MVSNet \cite{yao2019recurrent}, and multi-stage networks such as Point-MVSNet \cite{chen2019point}, Fast-MVSNet \cite{yu2020fast}, CVP-MVSNet \cite{yang2020cost} are included.
Furthermore, the current state-of-the-art unsupervised methods such as Unsup\_MVS \cite{khot2019learning}, M$^2$VS \cite{dai2019mvs2} and M$^3$VSNet \cite{huang2020m} are compared.

The quantitative results are shown in Table \ref{table1}.
From Table \ref{table1}, we can conclude that our proposed method outperforms previous unsupervised methods in all official metrics.
Furthermore, our proposed method can reconstruct better point cloud than traditional methods and some supervised methods in the metric of \emph{Overall}.
The supervised methods tend to have better performance in the metric of \emph{Accuracy}, while unsupervised methods usually achieve better performance in the metric of \emph{Completeness}.
The qualitative comparisons in Figure \ref{fig4} demonstrate that our proposed method is comparable with some of the SOTA supervised methods.

\noindent\textbf{Supervised vs Self-Supervised:}
From Table \ref{table1}, we can find that there still exists a clear gap of performance between SOTA supervised methods and previous unsupervised methods.
To provide a fair comparison without extra components, we compare our proposed self-supervision framework with supervised methods in the same network settings.
The only difference is that our model is trained without any ground truth depth maps.
The comparison is provided in Table \ref{table2}.
The supervised baselines are borrowed from previous papers(MVSNet from \cite{chen2019point}, CVP-MVSNet from \cite{yang2020cost}).
The results in Table \ref{table2} demonstrate that our proposed framework can compete on par with the supervised opponents in the same network settings.

\subsection{Ablation Studies}

\noindent\textbf{Effect of Different Prior Components:} 
To evaluate the effect of our proposed prior of semantic consistency and data augmentation consistency, we train the networks with different combinations of these self-supervised signals.
The quantitative results with different components in our proposed JDACS framework are summarized in Table \ref{table3} and Table \ref{table4}.
The model settings of JDACS in Table \ref{table3} and JDACS-MS in Table \ref{table4} is the same as the ones in Table \ref{table2}.
The qualitative visualization of the results of different components in JDACS-MS is provided in Figure \ref{fig5}.
The experimental results demonstrate that endowing these extra priors into the self-supervision training can promote the performance in MVS.
For example, as illustrated in Table \ref{table3}, the \emph{Overall} error metric decreases from 0.6777mm to 0.5953mm by including the prior of semantic consistency, from 0.6777mm to 0.5898mm with the help of involving data augmentation based branch.

\noindent\textbf{Effect of Semantic Cluster Numbers:}
Different from manual semantic annotations in supervised learning, the semantic concepts excavated in an unsupervised manner are ambiguous.
The number of semantic clusters $K$ is a significant hyper-parameter for determining the categories of common semantic concepts among different views.
Hence we conduct experiments about the effect of different semantic cluster numbers $K$ and the results are reported in Table \ref{table5}.
Furthermore, a brief visualization of these semantic clusters is provided in Figure \ref{fig6}.
From the visualization and the table, we can conclude that when the semantic clusters are more than 4, the localization of the semantic parts becomes less accurate than the ones with less than 4 clusters.
As a result, we select $K=4$ clusters as a default setting in our proposed method.

\begin{figure}[t]
\centering
\includegraphics[width=0.9\columnwidth]{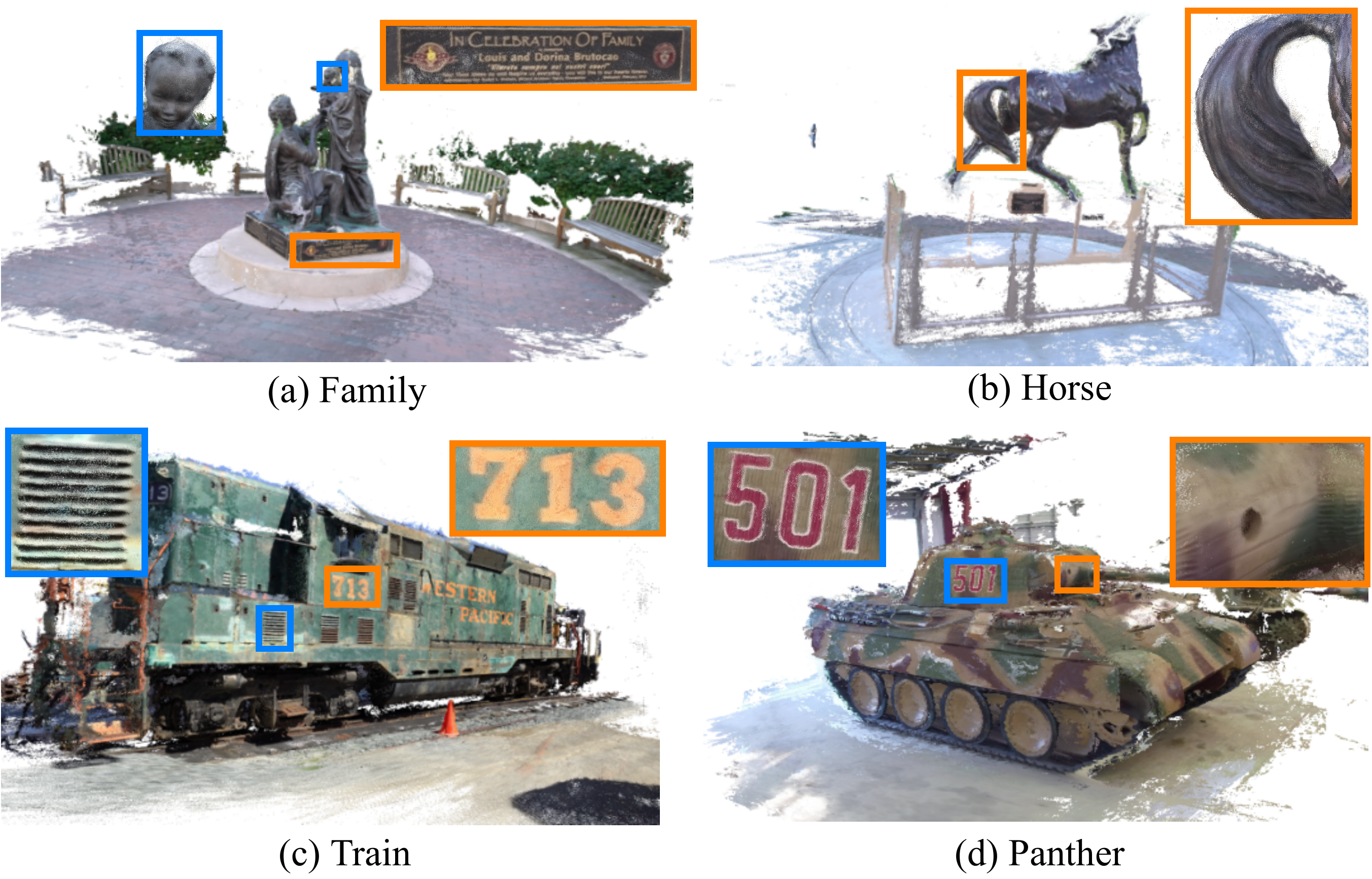} 
\caption{Visualization of the generated 3D point clouds without any finetuning on Tanks\&Temples dataset.}
\label{fig7}
\end{figure}

\subsection{Generalization}

In this section, we compare our proposed JDACS with previous unsupervised methods on Tanks\&Temples dataset.
Due to the requirement of more than 20G memories in GPU using the original post-processing tool provided by \cite{yao2018mvsnet}, instead, we use an open simplified version$\footnote{\url{https://github.com/xy-guo/MVSNet_pytorch}}$ which can be deployed on a GPU with 11G memories like RTX 2080Ti.
We follow the same hyper-parameter settings as MVS$^2$ \cite{dai2019mvs2}.
The quantitative comparison with previous unsupervised methods is provided in Table \ref{table6} and the visualization of the reconstructed dense point clouds is shown in Figure \ref{fig7}.
Our proposed JDACS has better performance by the mean score of 8 scenes than previous unsupervised methods, which is the best unsupervised MVS method until September 9, 2020.

\section{Conclusion}

In this paper, we have proposed a novel unsupervised learning based MVS framework, JDACS, aiming at alleviating the gap between supervision and self-supervision caused by the coarse hypothesis of color constancy.
On the one hand, our proposed method can enforce cross-view data-augmentation consistency into self-supervision with challenging variations.
On the other hand, we can excavate the implicit common semantic clusters among different views and enforce the cross-view semantic consistency to provide a semantic-level correspondence metric.
Experimental results on multiple benchmarks demonstrate the effectiveness of our proposed self-supervised framework.

\section{Acknowledgments}

This work was supported in part by the Shanghai Committee of Science and Technology, China (Grant No. 20DZ1100800), in part by the National Natural Science Foundation of China under Grant (61876176, U1713208), and in part by the Shenzhen Basic Research Program (CXB201104220032A), Guangzhou Research Program (201803010066). 
This work was done during his internship at Shenzhen Institutes of Advanced Technology, Chinese Academy of Sciences.

{\small
\bibliography{reference}
}

\clearpage

\section{Supplementary Materials}

\subsection{Supplementary Details for Implementation}

\subsubsection{Implementation of NMF}

NMF plays an important role in the Co-Segmentation branch of our JDACS framework. 
We use the multiplicative update rule to calculate the solution of NMF iteratively, as shown in Algorithm \ref{alg-nmf}.

\begin{algorithm}
\caption{Multiplicative Update Rule Based NMF}
\label{alg-nmf}
\begin{algorithmic}
    \STATE Set the number of segments as $K$; \\
    \STATE Set the number of maximum iterations as $\text{ite}_{max}$ and the tolerance constant as $\text{tol}$; \\
    \STATE Initialize non-negative matrices $P$ and $Q$ such that $P \geq 0, Q \geq 0$; 
    
    \FOR{each iterative step $v$, $1\le v \le \text{ite}_{max}$}
    
    \STATE $Q_{[i, j]}^{v+1} \leftarrow Q_{[i, j]}^{v} \frac{\left(\left(P^{v}\right)^{t} A\right)_{[i, j]}}{\left(\left(P^{v}\right)^{t} P^{v} Q^{v}\right)_{[i, j]}}$
    
    \STATE $P_{[i, j]}^{v+1} \leftarrow P_{[i, j]}^{v} \frac{\left(A\left(Q^{v+1}\right)^{t}\right)_{[i, j]}}{\left(P^{v} Q^{v+1}\left(Q^{v+1}\right)^{t}\right)_{[i, j]}}$
    
    \IF{$\| A - P^{v+1} Q^{v+1} \|_{F} \leq \text{tol}$}
        \STATE $P=P^{v+1}$, $Q=Q^{v+1}$, stop the iterative process
    \ENDIF
    
    \ENDFOR
\end{algorithmic}
\end{algorithm}

\subsubsection{Implemtation of JDACS-MS}

\begin{figure}[h]
\centering
\includegraphics[width=0.9\columnwidth]{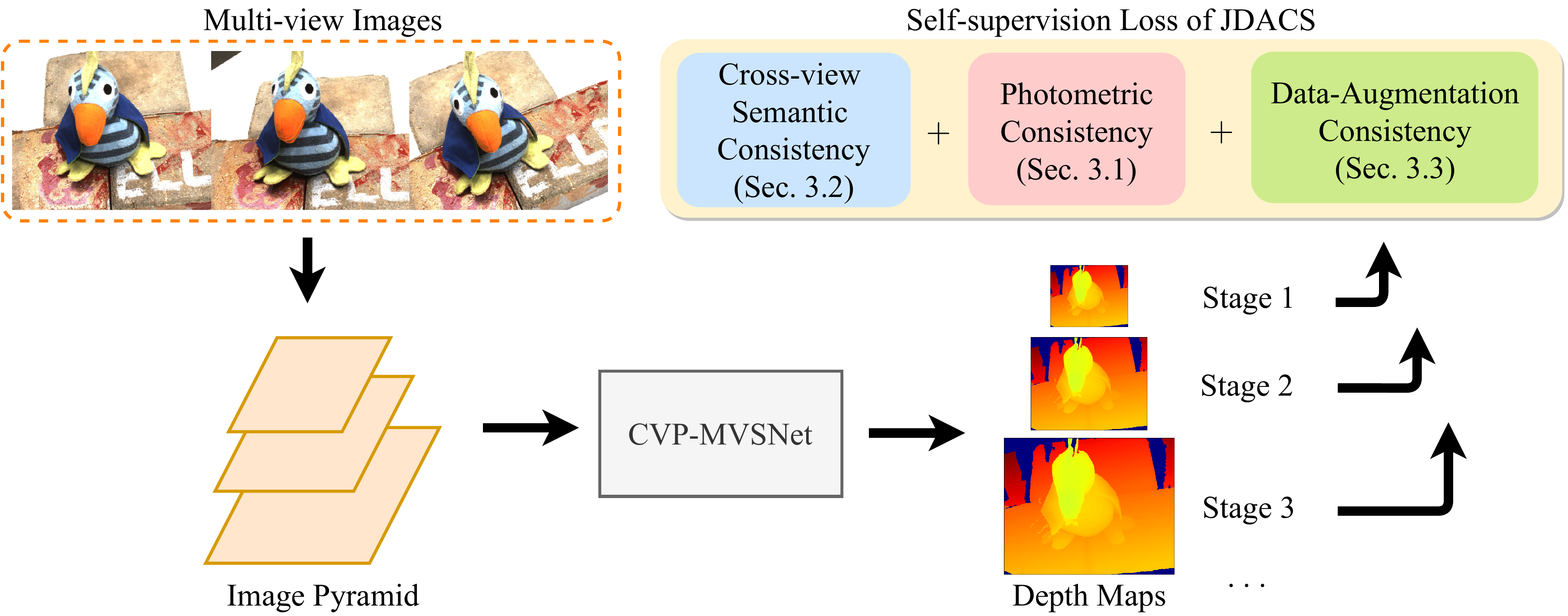} 
\caption{Brief illustration of JDACS-MS.}
\label{fig9}
\end{figure}

As mentioned in the main paper, if a multi-stage MVS-Net is applied in the Depth Estimation branch of JDACS, the framework is denoted as JDACS-MS. 
The predicted depth maps on all stages are utilized to calculate the self-supervision loss, as shown in Figure \ref{fig9}.
In default, we adopt CVP-MVSNet as the backbone network.

Similar to the loss function of JDACS (Equation \ref{eq8} in the main paper), the final objective of JDACS-MS can be constructed as follows:
\begin{equation}
\begin{split}
L_{J D A C S-M S}=\sum_{s=1}^{5}\left(\lambda_{1} L_{P C}^{s}+\lambda_{2} L_{S C}^{s}+\lambda_{3} L_{D A}^{s}\right.\\
\left.+\lambda_{4} L_{S S I M}+\lambda_{5} L_{S m o o t h}\right)
\end{split}
\label{eq9}
\end{equation}
where s represent each stage of the multi-stage MVSNet which is separated into 5 stages in default. The weights are empirically set as: $\lambda_1 = 0.8$, $\lambda_2 = 0.1$, $\lambda_3 = 0.1$, $\lambda_4 = 0.2$, $\lambda_5 = 0.0067$.

\subsubsection{Data-augmentation Consistency}

Various transformations are adopted for generating challenging samples, such as occluding mask, Gaussian noise, blur, random jitter in brightness, color and constrast. In JDACS with a backbone of single stage MVSNet, the input is the original multi-view images and differently randomized transformation is applied to each view. In JDACS-MS with a backbone of multiple stage MVSNet, the input is an image pyramid of multiple scales and different transformation is added to different level of the image pyramid on each view.

\subsubsection{How to Avoid Overflow in GPU Memory?}

The proposed framework possesses three parallel branches which may lead to an overflow in GPU memory during training.
It may be impracticable to train the model on a GPU with 11G memory, such as GTX 1080 Ti or RTX2080 Ti.
Hence, we conduct a simple trick to avoid the memory overflow by \emph{trading the GPU memory with time}.
In the training process, each step comprised of these three parallel branches and loss functions is separated into two sequential training step.
For example, we can propagate the Depth Estimation branch and Co-segmentation branch in the first step, and save the estimated depth map as pseudo depth label.
Then the Photometric consistency loss and Co-segmentation consistency loss are calculated, and the gradient is calculated during back-propagation.
After updating the gradients, the cached memories are cleared.
In the second step, the forward-propagation in the Data-augmentation branch is conducted and the weights are updated during the back-propagation.

\subsubsection{How to Adjust the Weights for Self-supervised Loss?}

In practice, the convergence is sensitive to the attribution of weights for each term in self-supervision loss.
If inappropriate weights are applied, it is likely to result in trivial solution in self-supervision.
Hence, it is important to balance the weights.
For the photometric consistency loss term, the weight is assigned following an open implementation\footnote{\url{https://github.com/tejaskhot/unsup_mvs}}.
For the co-segmentation consistency loss term, the weight is set according to the scale of the loss, which can be selected from 0.01 to 0.1.
For the data-augmentation consistency loss term, it is mentioned that the data-augmentation consistency is actually a strong regularization to the self-supervised framework.
In the starting phase of the training process, it may corrupt the convergence of self-supervision.
Hence, we set the weight of data-augmentation consistency loss to 0.01 as an initial value, and increase it by 2 times after each 2 epochs, acting as a warming up process.

\subsection{Visualization}

\begin{figure*}[ht]
\centering
\includegraphics[width=2.0\columnwidth]{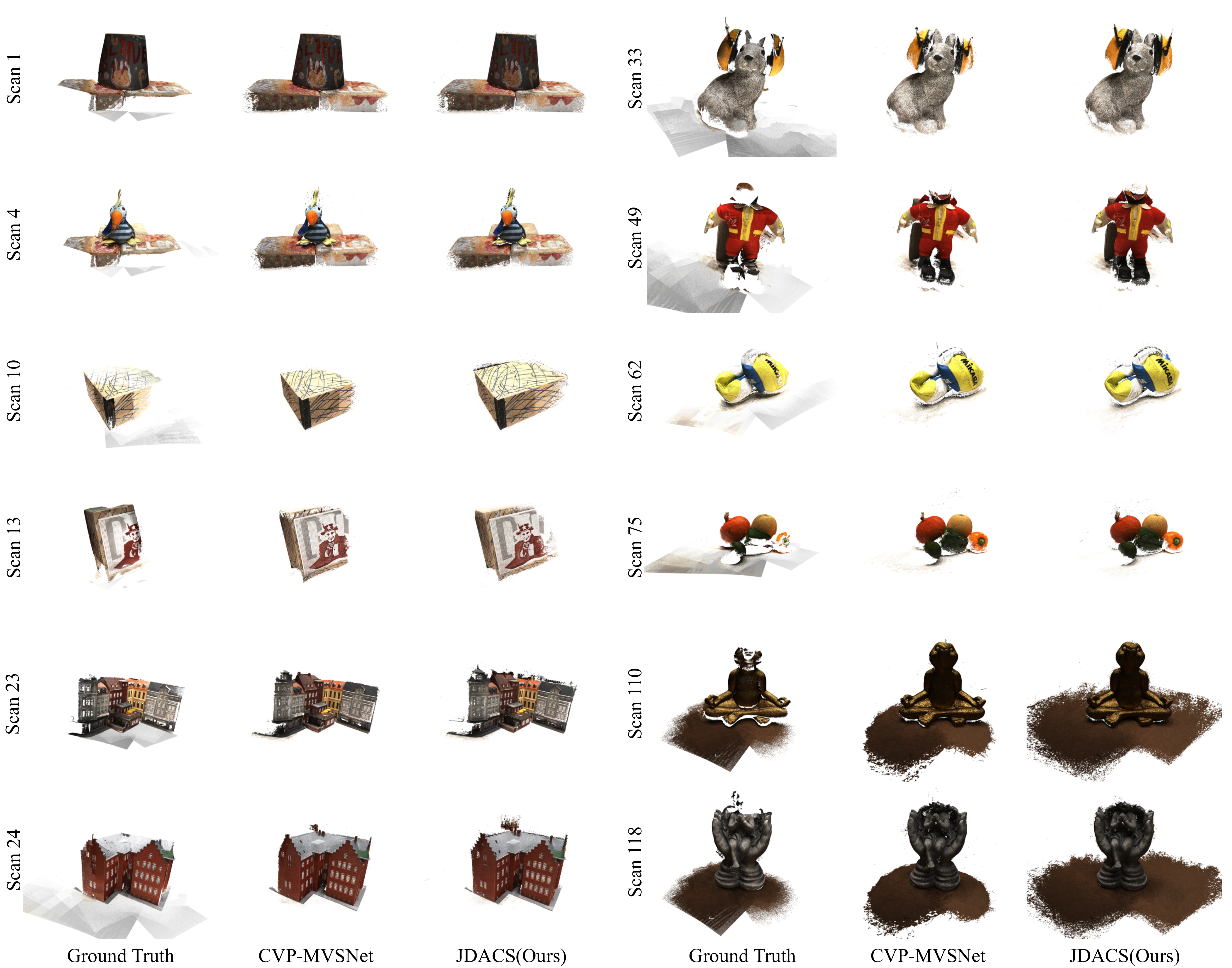} 
\caption{More qualitative comparison in 3D reconstruction between our JDACS and SOTA supervised method (CVP-MVSNet) on DTU dataset. From left to right: ground truth, results of CVP-MVSNet, our results.}
\label{fig10}
\end{figure*}

\begin{figure*}[ht]
\centering
\includegraphics[width=2.0\columnwidth]{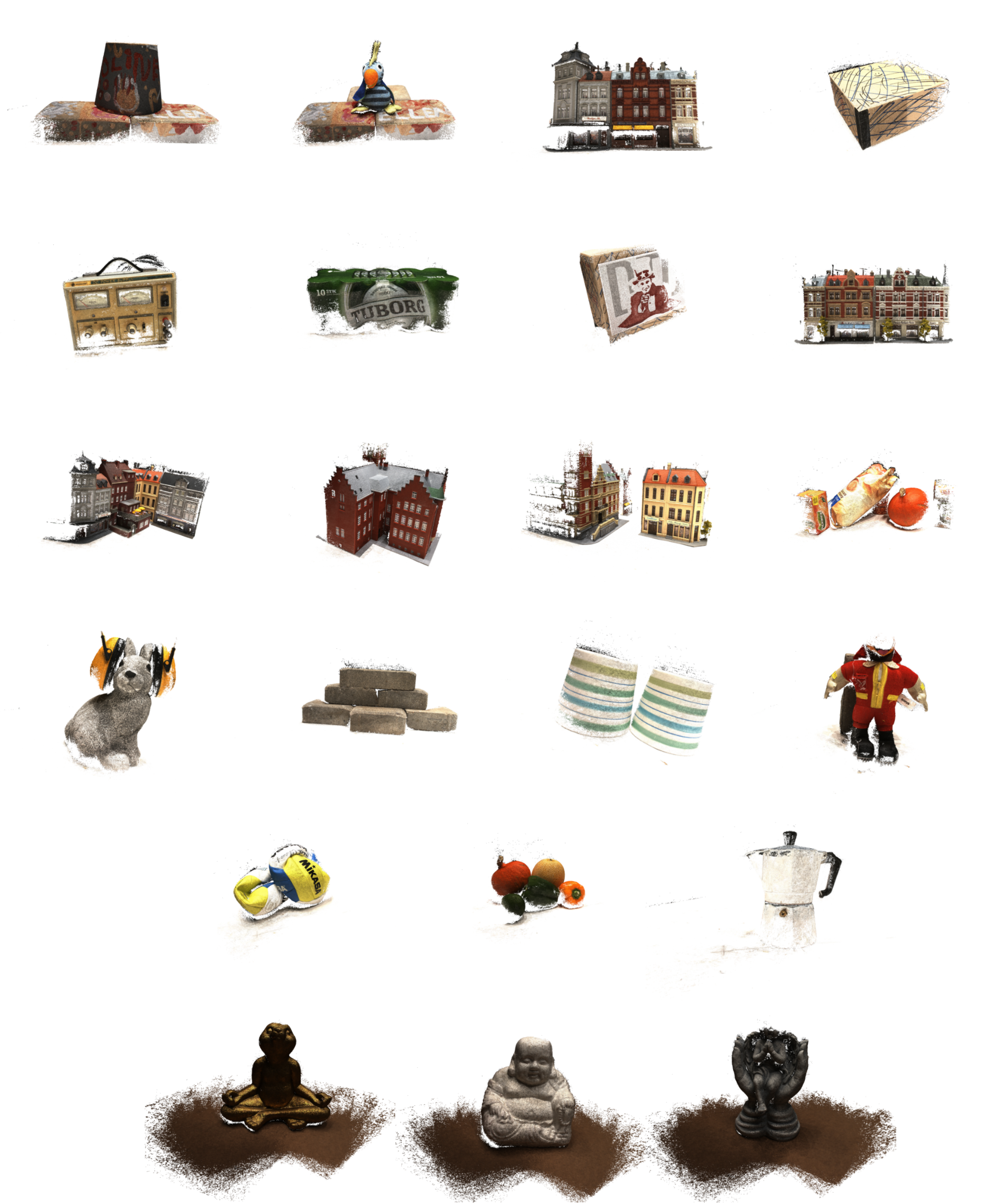} 
\caption{Point cloud reconstruction results on the DTU test set.}
\label{fig11}
\end{figure*}

\begin{figure*}[ht]
\centering
\includegraphics[width=2.0\columnwidth]{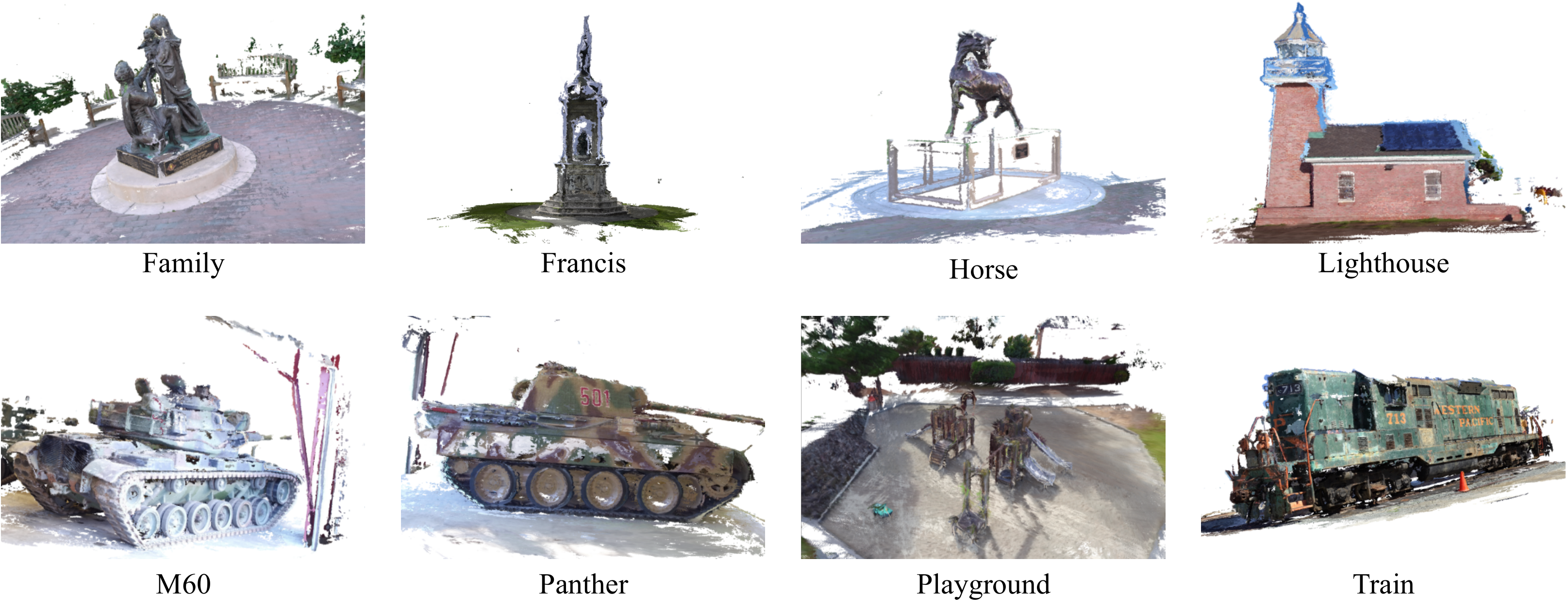} 
\caption{Point cloud reconstruction results on the Tanks\&Temples test set without fine-tuning.}
\label{fig12}
\end{figure*}

In addition to the qualitative comparison in 3D reconstruction shown in Figure \ref{fig5} of the main paper, we further provide more comparisons in Figure \ref{fig10}. Furthermore, we present the visualization of the reconstruction results on DTU dataset (Figure \ref{fig11}) and Tanks\&Temples dataset (Figure \ref{fig12}). Please refer to Table \ref{table1} and Table \ref{table6} in the main paper for quantative results on the datasets.

\subsection{Limitation and Discussion}

\subsubsection{Restriction of Coarse-grained Semantic Feature}

As shown in Table \ref{table5} and Figure \ref{fig7} in the main paper, the co-segmentation results can only provide coarse-grained semantic feature with no more than 4 semantic clusters. The reason is that the semantic centroids are clustered from the feature space of a pretrained VGG specialized for classification task, where only the coarse-grained semantics are enough to construct distinguishable clues. However, in intuition, fine-grained semantics can provide more effective priors of correspondence for self-supervision. In the future, more accurate and refined semantic features are required for further improving the performance of self-supervision.

\subsubsection{Restriction of Texture-less Region}

Although our proposed method can handle challenging cases with huge variation in color, it still fails to generalize to texture-less regions.
The convergence of all self-supervision reconstruction loss is only effective on colorful regions. 
Because any pixels in texture-less regions share the same color intensity, leading to the fact that self-supervision loss is fixed to 0 and becomes meaningless. 
However, the texture-less regions often appear in realistic scenarios, where self-supervision may be confused and fail to generalize. 
Exploration of handling texture-less regions may provide a potential direction in the future.

\end{document}